\documentclass[letterpaper, 10 pt, conference]{ieeeconf}

\IEEEoverridecommandlockouts

\usepackage{dblfloatfix}
\usepackage[noadjust]{cite}
\usepackage{amsmath,amssymb,amsfonts}
\usepackage{algorithmic}
\usepackage{graphicx}
\usepackage{textcomp}
\usepackage{xcolor}
\usepackage{pifont}
\usepackage{multicol}
\usepackage{multirow}
\usepackage{tabularx}
\usepackage[caption=false,font=footnotesize]{subfig}
\usepackage[citecolor=blue, colorlinks=true, urlcolor=black, linkcolor=red]{hyperref}
\usepackage{algorithm}
\usepackage{algorithmic}

\begin{document}

\title{\Large Structure from WiFi (SfW): RSSI-based Geometric Mapping of Indoor Environments}
\author{Junseo Kim*, 
        Jill Aghyourli Zalat*, 
        Yeganeh Bahoo*,  
        and Sajad Saeedi*
\thanks{
*Toronto Metropolitan University, Toronto, Canada\newline  \hspace*{1.4em}email: 
\{junseo.kim, jill.aghyourli, bahoo, s.saeedi\}@torontomu.ca
}
}

\maketitle

\begin{abstract}

With the rising prominence of WiFi in common spaces, efforts have been made in the robotics community to take advantage of this fact by incorporating WiFi signal measurements in indoor SLAM (Simultaneous Localization and Mapping) systems. SLAM is essential in a wide range of applications, especially in the control of autonomous robots. This paper describes recent work in the development of WiFi-based localization and addresses the challenges currently faced in achieving WiFi-based geometric mapping. Inspired by the field of research into k-visibility, this paper presents the concept of \emph{inverse k-visibility} and proposes a novel algorithm that allows robots to build a map of the free space of an unknown environment, essential for planning, navigation, and avoiding obstacles. Experiments performed in simulated and real-world environments demonstrate the effectiveness of the proposed algorithm.

\end{abstract}

\section{Introduction}
\label{sec:intro}

Simultaneous localization and mapping (SLAM) is a long-studied topic central to the control of autonomous robots. Much of what has been achieved thus far has focused on the use of exteroceptive sensors such as camera, laser-ranger, or ultrasound, which acquire information from the surrounding environment to generate map estimations~\cite{thrun2002robotic}. However, due to extraneous factors or physical limitations presented in some scenarios, 
these methods may not always be suitable. In such cases, one might turn to alternative methods that can attain similar mapping results. 

With the notable rise of WiFi networks becoming a staple in indoor buildings and public spaces, achieving SLAM using WiFi signal strength measurements, known as received signal strength indicator (RSSI), has become a prime focus in recent literature. Aside from being highly common, WiFi provides the advantage of being well-suited to environments where camera/laser/ultrasound sensors might fall short, such as in scenarios where there are privacy concerns or in badly illuminated environments~\cite{kudo2017utilizing}. 
Recent advancements in this approach, such as \emph{WiFiSLAM}~\cite{ferris2007wifi}, have mostly been focused on WiFi-based localization, or possibly estimating the position of the WiFi routers. 
It has also been shown that by relying on crowdsourcing~\cite{zhou2021crowdsourcing}, mapping using WiFi is possible, but this approach requires many Wi-Fi receivers. Additionally, RSSI, used in most of the algorithms, is known to be fluctuating and unreliable, particularly in dynamic environments~\cite{arun2022p2slam}. 
Therefore, building a complete WiFi-based SLAM solution remains a challenge, primarily because it is a difficult task to extract the geometric shape of the environment from WiFi data. 

\begin{figure}[t!]
  \centering
  \includegraphics[width=0.75\linewidth]{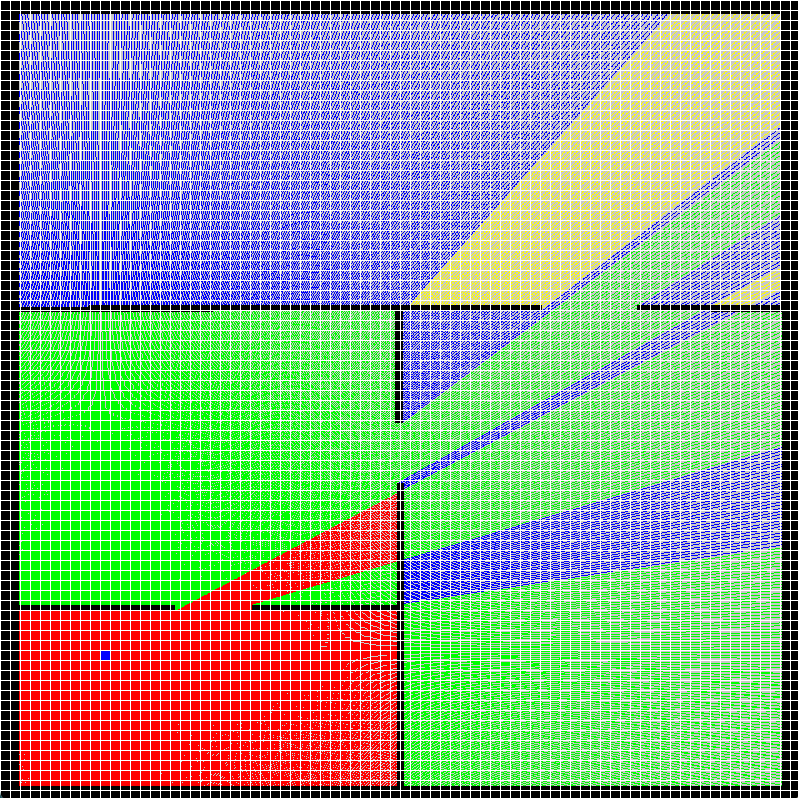}
  \caption{Demonstration of $k$-visibility where $k=0$ (red), $k=1$ (green), $k=2$ (blue) and $k=3$ (yellow) are shown. $k$-visibility refers to the number of times a signal from a reference point (e.g. a router, shown in dark blue) passes through a wall/obstacle when making a straight-line path to a desired location.}  
  \label{fig:k_visibility}
\vspace{-5 mm}
\end{figure}

In this paper, we use concepts from the field of {\it $k$-visibility}~\cite{o1987art}, as shown in Fig.~\ref{fig:k_visibility},  to devise a novel approach, coined Structure from WiFi (SfW), to generate a 2D geometric map of an indoor space using only WiFi signal-strength measurements and trajectory information. We propose the {\it inverse $k$-visibility} algorithm, which uses probabilistic modeling of known $k$-visibility information to estimate an explicit model of the environment.

The main contribution of the work is bringing $k$-visibility concepts into robotics mapping problems and proposing a mapping algorithm that maps most of the free space using WiFi RSSI signals without relying on sensors such as lidar, radar, or camera. Mapping free space is significant, as it allows the robots to plan paths without colliding with obstacles, essential for the control of many autonomous systems. 
Evaluation of the work in simulation and real-world settings demonstrates the significance of the method. See the videos on the website of the project\footnote{\href{https://sites.google.com/view/structure-from-wifi/home}{https://sites.google.com/view/structure-from-wifi/home}}.

The rest of this work is organized as follows: 
Sec.~\ref{sec:related} presents the literature review. 
Sec.~\ref{sec:background} describes the background material.
Sec.~\ref{sec:dense} proposes the inverse $k$-visibility algorithm.
%
Sec.~\ref{sec:sparse} extends the inverse $k$-visibility algorithm to real-world situations where sparse information is available.
Sec.~\ref{sec:exp} presents the experimental results.
Finally, Sec.~\ref{sec:con} discusses future works and concludes the paper.

\section{{Literature Review}}\label{sec:related}
The following section summarizes recent advances in WiFi-only SLAM systems and wireless-based mapping, making note of the difficulties being faced in creating an explicit map of the environment using signal-strength measurements. 

The utilization of WiFi-based systems to achieve localization has seen widespread adoption in recent years. Ferris {\it et al.}~\cite{ferris2007wifi} achieved localization by mapping high-dimensional signal strength measurements into a 2D latent space, solving SLAM using a technique known as Gaussian process latent variable modeling (GP-LVM). Other works adopt a similar approach, including \cite{xiong2017diversified, miyagusuku2016improving}. Graph-based approaches are also used for SLAM, such as~\cite{huang2011efficient, liu2019collaborative, herranz2016wifi, arun2022p2slam}. Improving WiFi observation models can also lead to improved mappings, such as the works described in \cite{kudo2017utilizing, he2015wi}.

Several works have taken advantage of sensors inherent to smartphones and have attempted to infer the floor plan of an indoor environment via a crowdsensing approach~\cite{zhou2021crowdsourcing}. Much of the research in this field has approached mapping using inertial sensors supported by WiFi signal strength measurements, both of which are readily available in commercial smartphones. Recent papers include~\cite{luo2014piloc, shin2011unsupervised, zhou2015alimc, zhou2018graph, shen2013walkie, jiang2013hallway, alzantot2012crowdinside, liang2016sensewit}. The benefit of these works is limited to the fact that they only achieve the creation of ``traversable maps", which can only depict traversable areas in the environment, whereas occupied areas remain unknown. All such works mentioned have relied on crowdsourced and/or public data, which are not always available and thus limit the scope of application.  

With directional antennas, Gonzalez-Ruiz and Mostofi~\cite{gonzalez2013cooperative} presented a framework for creating a non-invasive occupancy grid map using wireless measurements, achieving a 2D representation of an environment using a coordinated robot setup, where information is obtained through walls or other obstacles. Other works looking into the closely-related field of Ultra Wide-band (UWB) SLAM have used similar directional antenna techniques based on UWB signal path propagation modeling~\cite{deissler2010uwb, deissler2012infrastructureless}.

\section{Background: $k$-visibility}\label{sec:background}

Here, the concept of {\it $k$-visibility}, which shall be extensively used in the following work, will be introduced as an alternative perspective of WiFi systems.

{The field of research into \emph{$k$-visibility}, first introduced as the {\it modem illumination problem}~\cite{fabila2009modem}, concerns itself with WiFi-based systems from the perspective of finding the amount of area that can be visualized from router points located on the vertices of a polygon. It is an extension of the problem of \emph{visibility}, which attempts to find the region visible to a point in a polygon~\cite{o1987art}. Two points \emph{p} and \emph{q} on a simple polygon \emph{P} are said to be mutually visible if the line segment joining \emph{p} and \emph{q} does not cross the exterior of \emph{P}~\cite{o1987art}. The visibility concept arose from the Art Gallery Problem proposed by Victor Klee in 1973~\cite{klee1969every}, which seeks to find the number of guards sufficient to cover the interior of an \emph{n}-wall art gallery room. The region visible to a certain point is a \emph{visibility polygon}~\cite{o1987art}. On the other hand, \emph{p} and \emph{q} are said to be \emph{$k$-visible} to one another if the line segment joining them crosses the exterior of \emph{P} at most \emph{k} times~\cite{o1987art}. 
The $k$-visible region can then be found for a given value of \emph{k} from a certain vantage point \emph{p}. 
Fig.~\ref{fig:k_visibility} shows a map with various \emph{$k$-visibility} values from the perspective of a point highlighted in dark blue (inside the red region). All red cells are 0-visible, green cells are 1-visible, blue cells are 2-visible, and yellow cells are 3-visible.
The vantage point \emph{p} is denoted as a \emph{k-transmitter}, and all $k$-visible polygons can be found for every value $\{0, 1, .., k\}$~\cite{o1987art}. 
%
Various methods exist for determining $k$-visibility plots with different complexity level~\cite{bajuelos2012hybrid, bahoo2020computing, bahoo2019time}. 
}

\section{{Dense Inverse $k$-visibility}} \label{sec:dense}

In this section, we propose the \emph{dense inverse $k$-visibility} concept. The main principle behind the algorithm is first explained. Then, in the next section, we extend this concept to sparse $k$-visibility with its application in robotics.

\begin{figure}
  \centering
  \includegraphics[width=0.75\linewidth]{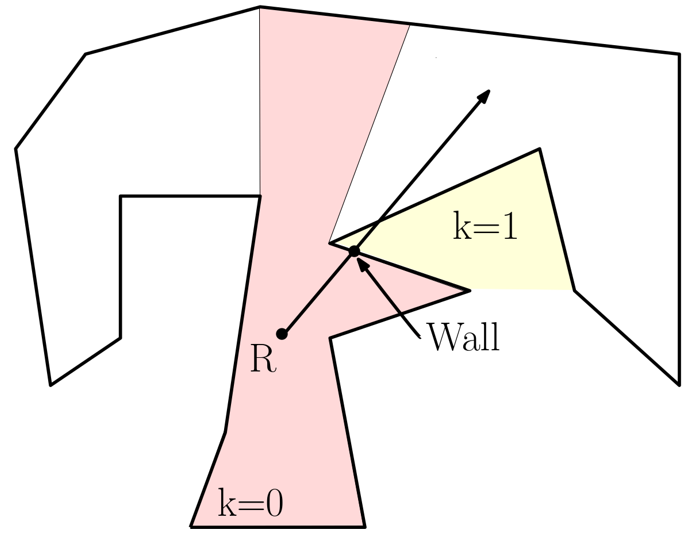}
  \caption{Diagram demonstrating the ray-drawing principle upon which the inverse $k$-visibility algorithm is based. 
  The wall is located at the exact coincidence of two consecutive $k$-value regions along a ray cast from the router point. Note that only the region of interest in the $k=1$ region is shown.}  
  \label{fig:raydrawingprinciple}
\end{figure}

In $k$-visibility related works, the goal is to assign $k$-values from a known $k$-transmitter, or router point, by casting rays in various directions. With the map (represented as a simple polygon) being known, one looks to obstructions encountered along each ray being cast from the router point and assigns $k$-value regions based on these obstructions. It can be observed that where there are consecutive $k$-value regions along a ray cast from the transmitter point, a wall is located at the coincident point between these two regions. The inverse $k$-visibility algorithm is based upon this concept: instead of having a map of the environment as an input, one uses known $k$-value regions to infer the map by casting rays from the router point and observing for consecutive $k$-value regions along the ray. Where these are found, a wall location can be marked. Fig.~\ref{fig:raydrawingprinciple} demonstrates this ray drawing principle. 

If the full $k$-visibility plot is given, one can easily see the border between consecutive $k$-value regions. This information can be used in a ``full" inverse $k$-visibility algorithm to generate a map of the environment. Such an algorithm would take as an input a colour map of all superimposed $k$-visibility plots for all values $k$ corresponding to a given environment and router point location, as in Fig.~\ref{fig:k_visibility}. It can be seen that walls are located on the direct border between consecutive $k_{i-1}k_i$ portions of the $k$-visible map. The full inverse $k$-visibility algorithm functions by detecting shared pixels among these successive $k$-value areas and writing them to a separate image, creating a black-and-white outline of the actual environment.


\section{Sparse Inverse $k$-visibility}\label{sec:sparse}
In practical applications, determining the full $k$-visibility plot for every $k$ value is not feasible. Thus, we explore the partial recreation of an environment using sparse $k$ values, an approach we call {\it sparse inverse $k$-visibility}. Similar to the full inverse $k$-visibility algorithm, sparse inverse $k$-visibility involves relating coordinates of known consecutive $k$ value coordinates located along the same ray cast from the router point, whose location is assumed to be known. The locations can be determined via dead reckoning using inertial measurement units and wheel encoders (if available on mobile robots). 

\begin{figure}
  \centering
  \captionsetup{justification=centering}
  \includegraphics[width=.9\linewidth]{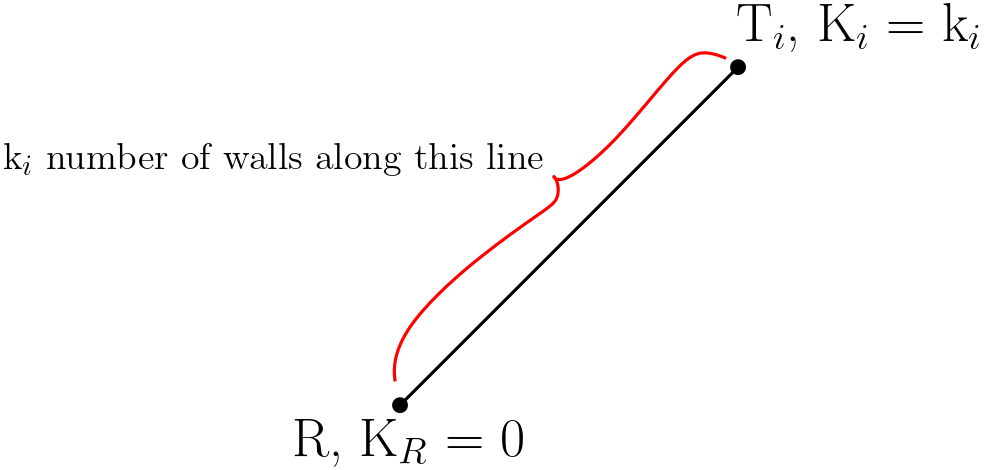}
  \caption{Ray-drawing for an arbitrary trajectory coordinate $T_i$ which has an associated $k$-value $k_i$. By definition of $k$-visibility, the ray $\overline{RT_i}$ must have a $k_i$ number of walls along the ray.}
  \label{fig:rti_line_drawing}
\end{figure}

Here, an overview of the algorithm is described. Consider an indoor environment whose router point location is known; the coordinates are denoted as $R = (r_x, r_y)$. A user or mobile robot walks along a path throughout this environment, moving from room to room. Assume that the set of $(x,y)$ coordinates of the trajectory taken across the map can also be known, and let $T = [T_1, T_2, ..., T_{max}]$ describe this set of coordinates, where $T_i$ represents the $(x,y)$ position of a point along the $i^{th}$ coordinate in the trajectory array. Assume further that the set $K$ of $k-$values corresponding to each pair of trajectory coordinates can also be known, where $k$ is an integer representing the number of obstacles that lie between a given point on the trajectory $T_i$ and the router $R$. \\
\indent The \emph{probabilistic sparse inverse $k$-visibility} algorithm then works in three parts:
\begin{enumerate}
    \item Extracting $k$-values: The trajectory is sub-divided according to associated $k$-values to coordinates (See Sec.~\ref{subsec:k});
    \item Mapping Free Space: Free space is determined  (See Sec.~\ref{subsec:free});
    \item Mapping Occupied Space: the occupied pixels are probabilistically determined using a three-step process of ray drawing, ray segmentation, and Gaussian probability assignment (See Sec.~\ref{subsec:occupied}).
\end{enumerate}
The algorithm outlined herein is based on the principle of drawing rays from $R$ to a coordinate being analyzed along the trajectory $T_i$, which has a corresponding $k$-value $k_i$. The environment is analyzed in a grid map format, emulating an occupancy grid map style of mapping. 
In an occupancy grid map, every pixel (cell) is either unknown, free, or occupied. See~\cite{thrun2002robotic} for more information.
Based on the principles of $k$-visibility, the ray $\overline{RT_i}$ is bound to have a $k_i$ number of walls along this line. Walls are taken to be distinct cells in the grid map that lie along the $\overline{RT_i}$ line. This concept is shown in Fig~\ref{fig:rti_line_drawing}. 

\subsection{Extracting $k$-values}\label{subsec:k}

The work done by Fafoutis {\it et al.}~\cite{fafoutis2015rssi} in creating a wall prediction model allows for the quantifying of $k$-values in an experimental scenario corresponding to the trajectory plotted. The authors sought to use RSSI measurements to predict the number of walls between a wearable sensor and an access point. The RSSI-based wall prediction function transforms an RSSI measurement at a given point, $P_{RSSI}$, to the predicted number of walls to the access point for an upper limit of $K$ number of walls based on a sequence of RSSI thresholds, $t_{1}, t_{2}, ..., t_{K}.$ The function is as follows~\cite{fafoutis2015rssi}. 
\begin{equation} 
f(P_{RSSI}) = \begin{cases} 
      0 & P_{RSSI} > t_{1} \\
      1 & t_{1}\geq P_{RSSI} > t_{2} \\
      ... \\
      K-1 & t_{K-1}\geq P_{RSSI} > t_{K} \\
      K & t_{K}\geq P_{RSSI}
   \end{cases}
\end{equation} 
Additionally, the RSSI signal cannot be amplified through a wall. Thus, the authors show the following relation:
\begin{equation}
    t_{K} < t_{K-1}: k\in [1, K]
\end{equation} 
For the purposes of Structure from Wifi mapping, the K-Means algorithm presented by the authors is most suitable, given that it is unsupervised and does not require labeled data. 

For an upper limit of $K$ walls, the K-Means algorithm presented by Fafoutis {\it et al.} outputs $K+1$ centroids, sorted in descending order: $C_{0}, C_{1}, ..., C_{K}$. Thus, the authors define the RSSI thresholds as follows.
\begin{equation}
    t_{K} = \frac{C_{k-1} + C_{k}}{2}: k\in [1, K]
\end{equation}
We use these equations in conducting an experimental trial of the sparse inverse $k$-visibility algorithm.

In the real world, RSSI signals fluctuate and are noisy. To reduce the noise of the RSSI values and eliminate the RSSI fluctuations, a sliding window filter is applied to the RSSI values along the trajectory.

\subsection{Mapping Free Space}\label{subsec:free}

\color{black}

\begin{figure}
  \centering
  \captionsetup{justification=centering}
  \includegraphics[width=0.75\linewidth]{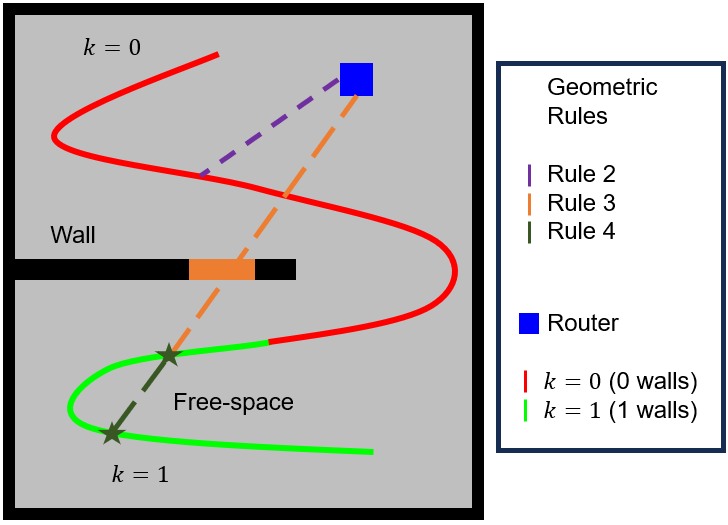}
\vspace{-2 mm}
  \caption{Visual demonstration of the geometric rules with $k=0$ and $k=1$ area. Rule 1 was excluded from the legend for clear visualization as it shows the trajectory of the robot.}
  \label{fig:rules}
\end{figure}

To map the free space, the following geometric rules, as shown in Fig.~\ref{fig:rules}, are utilized. We start with the assumption that the pixels of the map are unknown. The trajectory is first segmented based on the $k$-values, as described in Sec.~\ref{subsec:k}. Then, the following rules are implemented to determine the free space: 

\color{black}


    \noindent {\bf Rule 1}: The robot's trajectory is considered to be free unless it hits an obstacle, which can be detected by contact sensors.
    
    \noindent {\bf Rule 2}:: If $k=0$, pixels on the line segment between the router and the robot are free space.
    
    \noindent {\bf Rule 3}:: If $k \geq 1$, there are walls between the router and the robot.
    
    \noindent {\bf Rule 4}:: for any line emanating from a router, if the line intersects the robot trajectory at two points with the same $k$-values, the pixel residing on the line that is between the two points belongs to the free space.

\subsection{Mapping Occupied Space}\label{subsec:occupied}

\begin{figure}
  \centering
  \captionsetup{justification=centering}
  \includegraphics[width=1\linewidth]{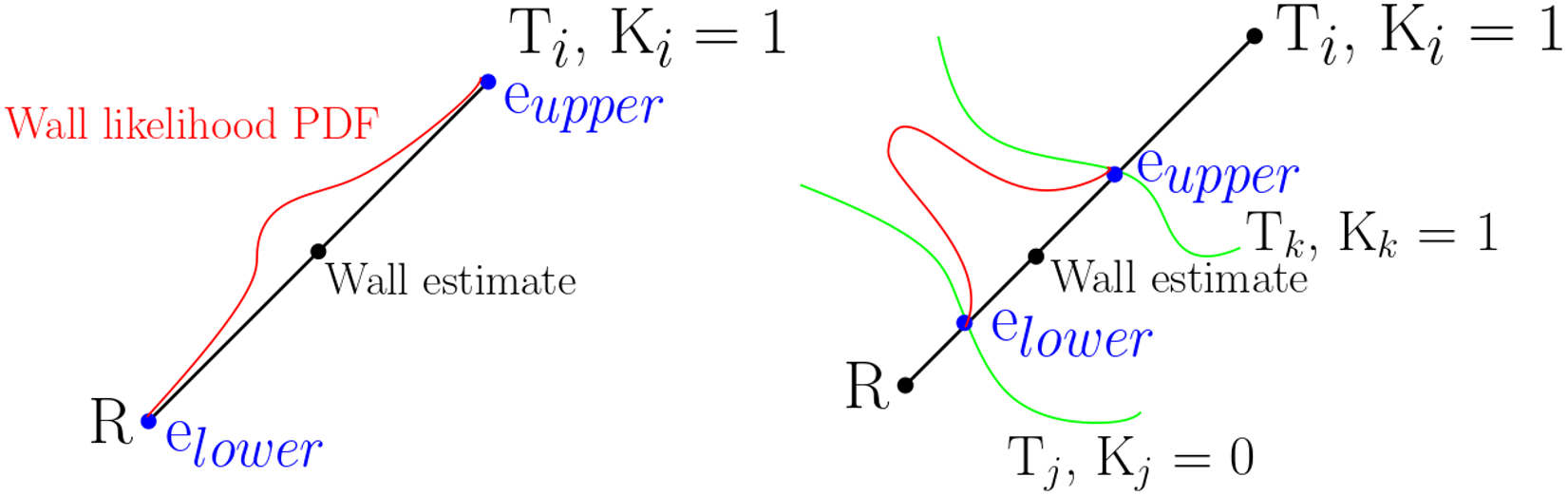}
\vspace{-5 mm}
  \caption{Initial wall estimate along a ray (left). Improved wall estimate along a ray after updating the lower and upper endpoints (right).}
  \label{fig:gaussdemo12}
\end{figure}

Similar to the free space, segmenting the trajectory based on $k$-value and continuity is the first step in the algorithm for finding occupied space. The algorithm groups trajectory coordinates according to their corresponding $k$-values. 
%
%
A probabilistic model is then applied to identify wall and obstacles. Consider the ray $\overline{RT_i}$ drawn in Fig~\ref{fig:gaussdemo12}, where the associated $k$-value for the trajectory position $K_i = 1$. By definition of $k$-visibility, it is known that there exists $K_i$ number of wall cells along  $\overline{RT_i}$, where, in this case, exactly one wall cell will lie on the line. With no other known information, an assumption one can make about the location of this wall cell is that it is situated at the midpoint of  $\overline{RT_i}$. The longer the length of the ray, the less reasonable this assumption becomes. 
We represent this certainty probabilistically as: 
\begin{equation} \label{intermediate_cell_probability}
    \mu_j = \frac{e^{-(\frac{1}{M})^2}d_j}{L},
\end{equation}
\\ where $\mu_j$ is the probability of the $j^{th}$ cell in the ray being a wall, $M$ is the number of intermediate cells along the ray, $L$ is the length of the ray, and $d_j$ is the distance from the $j^{th}$ cell to the midpoint of the ray. The intermediate cells of a ray refer to all cells along the ray, excluding the endpoints.
\\ The prediction made by drawing ray $\overline{RT_i}$ in Fig.~\ref{fig:gaussdemo12} may be improved by updating the ray endpoints such that the result is a shorter line, and consequently a smaller number of intermediate cells, which improves the probability estimate. 
To start with, the lower endpoint of a ray, denoted as $e_{lower}$, is the router point $R$. Similarly, the upper endpoint, denoted as $e_{upper}$, is the trajectory point $T_i$. One notices that, in some instances, the ray may cross other points along the trajectory. These serve as the updated endpoints of the ray $\overline{RT_i}$, as seen in Fig.~\ref{fig:gaussdemo12}. The lower endpoint is updated if the other trajectory point encountered $T_j$ has $K_j = 0$; otherwise, the upper endpoint is updated if $K_j \geq 1$.

\begin{figure}
  \centering
  \captionsetup{justification=centering}
  \includegraphics[width=0.75\linewidth]{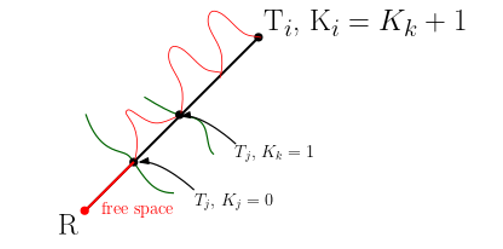}
\vspace{-2 mm}
  \caption{Assigning of probability distributions based on differences in $k$-values among ray subsegments.}
  \label{fig:raysubsegs}
\end{figure}

The above can be generalized for an algorithm that analyzes $k$-values that may be greater than 1. For every trajectory coordinate, a ray $\overline{RT_i}$ is drawn, and the trajectory crossings are determined as above. The ray is then divided into subsegments based on the trajectory crossings encountered. The difference $\Delta k$ is then determined between the endpoints of every subsegment, as in Fig~\ref{fig:raysubsegs}. If $\Delta k = 0$, that is, the two endpoints are equal in $k$-value, then the area is assigned to be free space. If $\Delta k = 1$, then there must lie exactly one wall along this subsegment, and an unimodal probability distribution is assigned, with the wall likely to be at the midpoint given no other information. Lastly, if  $\Delta k > 1$, more than one wall lies along this line, and a multimodal distribution is assigned to this segment.

With rays being drawn for every cell along the trajectory, some rays may ``see" a cell that another ray has already seen. In this instance, the cell would have already been assigned a probability based on Eq. \ref{intermediate_cell_probability}. The probability of the cell is then updated by combining the probabilities from the prior ray drawn and the current ray based on each estimation's associated uncertainty.
\begin{equation} \label{intermediate_cell_probability}
    \mu = \frac{\sigma_1^2}{\sigma_1^2+\sigma_2^2} \mu_2 +  \frac{\sigma_2^2}{\sigma_1^2+\sigma_2^2} \mu_1
\end{equation}
where $\mu$ is the combined probability, $\mu_1$ and $\mu_2$ are the probabilities of the prior ray and the current ray, respectively, and $\sigma_1$ and $\sigma_2$ are the uncertainties associated with the prior ray probability and the current ray probability, respectively. \color{black} All these steps are summarized in Alg.~\ref{alg:sfw_algorithm}. \color{black}

\begin{algorithm}[t!]
\color{black}
\caption{Mapping Free Space}
\label{alg:sfw_algorithm}
\begin{algorithmic}[1]
\STATE \textbf{initialize:} Robot position $(x, y) \in T$, 
\STATE trajectory $T \in M$, map $M_{ij}\in\{\mathbb{R}^2\}$ $\forall i, j, M_{ij} \gets 127$, 
\STATE and router position $(a,b) \in M$.

\WHILE{Robot goes to next position $(x,y)_i \gets (x,y)_{i+1}$}
    \STATE Set $M_{(x,y)_{i}} \gets 0$ (free space) [Rule 1].
    \FORALL{\((t_x, t_y)\in M \) from \((x, y)\) to \((a, b)\)}
        \STATE \( L_i \gets \text{$P_{RSSI}$ from } (t_x, t_y) \text{ where } L_i, P_{RSSI}\in\mathbb{R}\).
        \STATE \( K_i \gets \text{$f(P_{RSSI})$ from } (t_x, t_y) \text{ where } K_i\in\mathbb{R}\).

        \IF{$K_i = 0$,} 
            \STATE $L_i = L_i + \sigma$ (increase prob. free space) [Rule 2].
        \ELSIF{$K_i \geq 1$,}
            \STATE $L_i = L_i - \sigma$ (increase prob. of walls) [Rule 3].
        \ENDIF
        \IF{$\exists c,d \in\mathbb{R} \text{ such that } K_c=K_d \text{ and } c\leq d$,}
            \FORALL{$L_i \text{ from } L_c \text{ to } L_d$}
                \STATE  $L_i=L_i + \sigma$ (increase prob. free space)
                \STATE [Rule 4].
            \ENDFOR
        \ENDIF

        \STATE Update map $M$ by \( P_{RSSI} \text{ of } (t_x, t_y) \gets L_i\).
    \ENDFOR
\ENDWHILE
\end{algorithmic}
\end{algorithm}

\color{black}


\section{Experimental Results} \label{sec:exp}
{The following section demonstrates the simulated and real-world experiment, showcasing the implementation of sparse inverse $k$-visibility.

\begin{figure}
  \centering
  \captionsetup{justification=centering}
  \includegraphics[width=1\linewidth]{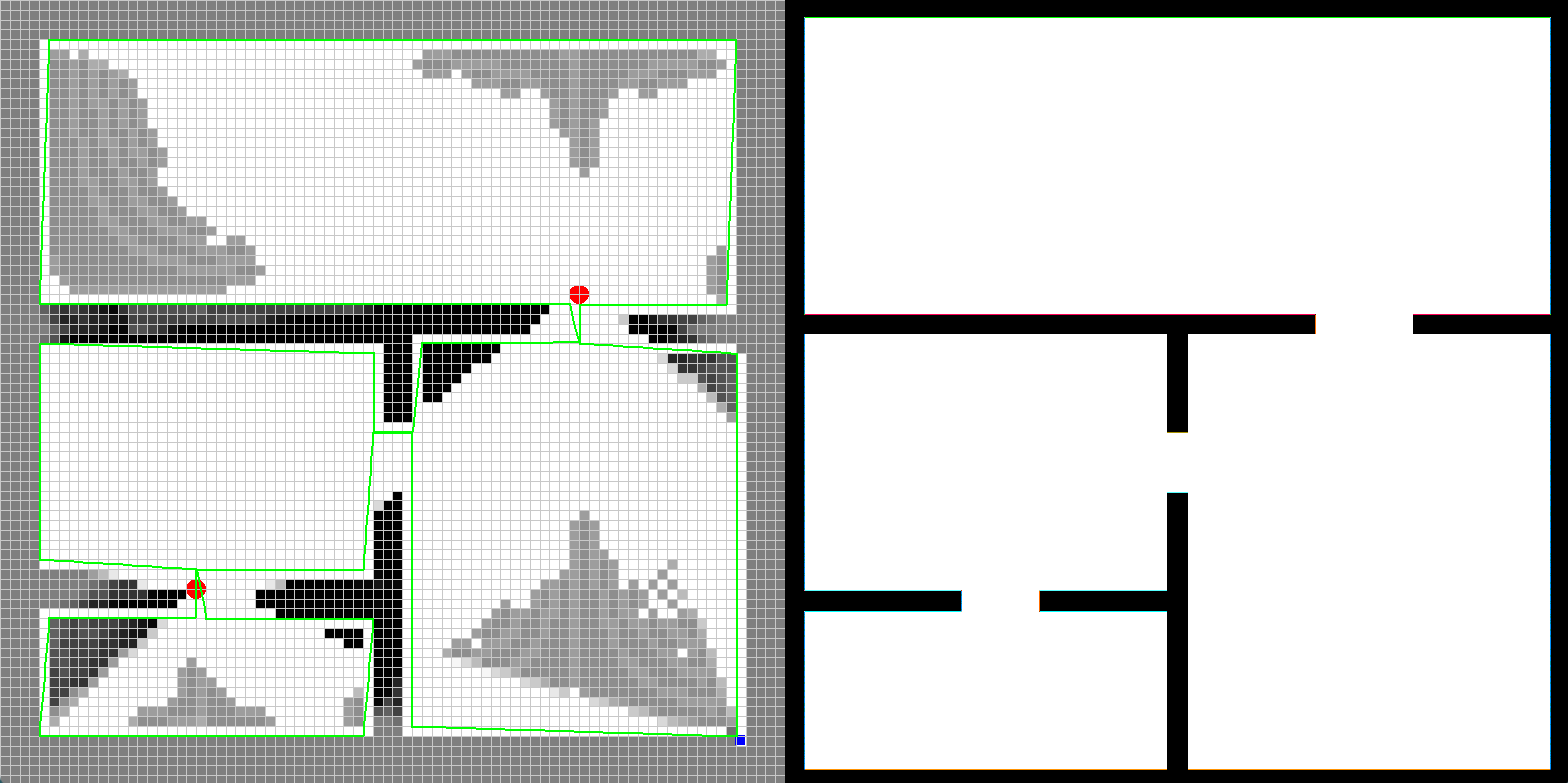}
\vspace{-7 mm}
  \caption{Simulated results based on the proposed algorithm. (left): free space and occupied cells are shown, with the trajectory of the robot in green. (right): Ground-truth map~\cite{"li2019houseexpo"}.}
  \label{fig:simgroundtruth}
\end{figure}

A demonstration of the algorithm was carried out on a map provided by the HouseExpo dataset~\cite{"li2019houseexpo"}. The ground truth and results are shown in Fig.~\ref{fig:simgroundtruth}. In this experiment, an idealized trajectory is drawn around the walls of each room, and every trajectory coordinate is associated with a $k$-value according to its position from the router. One can observe that free space and walls are approximated in roughly the same areas as the walls in the ground-truth.

For the real-world experiments, the ground-truth data was collected using a $360$ Laser Distance Sensor (LDS-01), and the 2D occupancy grid map was built using Gmapping~\cite{grisetti2007improved}.

The SfW algorithm's results largely depend on the quality of the trajectory obtained. Odometry data from the TurtleBot3 was used to obtain $x$ and $y$ coordinates of the robot for all the points on its trajectory. WiFi signal strength measurements were collected via the terminal to collect signal strength measurements. We did two different trajectories at two different locations, demonstrating the effectiveness of our proposed method in different environments. We assumed the location of the router is known, though if the location is unknown, it can be calculated using other algorithms easily.

\begin{figure}
  \centering
  \captionsetup{justification=centering}
  \includegraphics[width=1\linewidth]{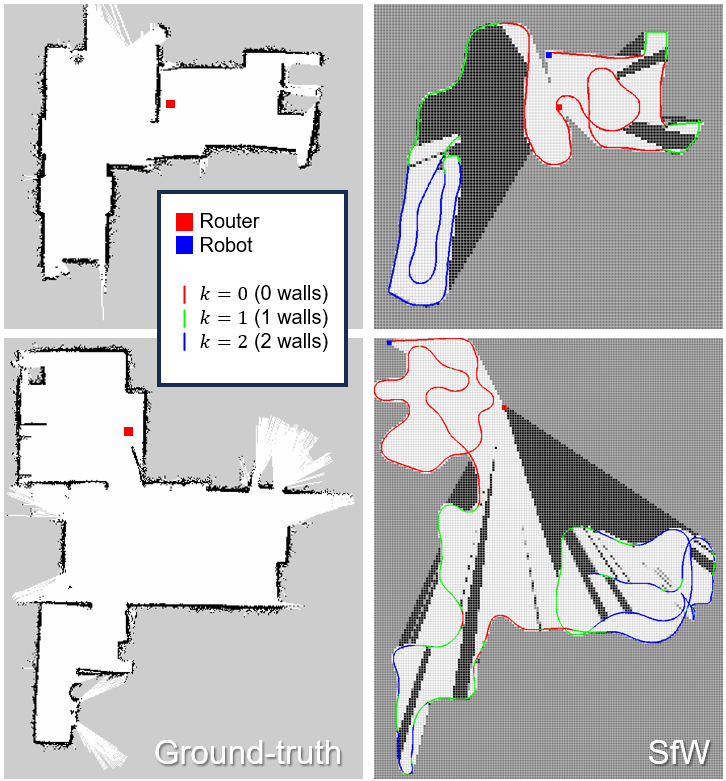}
\vspace{-7 mm}
  \caption{Real-world experimental results: Experiment 1 (top) and Experiment 2 (bottom), showing the ground-truth map using Gmapping (left) and the proposed WiFi-based map (right). Our proposed method with RSSI-based mapping detects most of the free space and  potential walls with $k\geq 1$.}
  \label{fig:data4and2fig}
\end{figure}

The first experiment was performed in a location composed of three small rooms with a total area of $15.7~m^2$. Fig.~\ref{fig:data4and2fig} (top left) shows the flood-plan of the environment. The robot was driven in the environment, collecting 3159 RSSI data points. As shown in Fig.~\ref{fig:data4and2fig} (top right), $k$ values based on the RSSI signal generally captured the free space and potential wall locations. 
The part of the trajectory with $k=0$ is highlighted in red, $k=1$ in green, and $k=2$ in blue. There were no higher $k$-values in this experiment.

Due to fluctuations by the RSSI signals and the distance loss, it failed to correctly detect the area with a $k$ value of zero and where the distance was very close to the router. Such trajectories were typically challenging when using RSSI signals due to the difference between the wall loss and the distance loss. 

The second experiment covered a longer trajectory with 4713 RSSI data points in an area of $34.2~m^2$. This is shown in Fig.~\ref{fig:data4and2fig} (bottom row). We assessed that our proposed method could estimate most of the free space. We are confident that utilizing the more optimized RSSI signal would have led to a higher estimation of the $k$ value as well as the wall prediction.

In both experiments, the robot obtained knowledge of the free space without exploring all of the free space and without using sensors such as a camera, radar, or lidar. This is significant, as it can be used in applications where conventional camera/radar/lidar is not available.
}

\section{Future work and Conclusions} \label{sec:con}
In this paper, we presented a promising research direction for WiFi-based geometric mapping, which is able to indicate the approximate structure of indoor environments, notably the free space. The free space is important for the robot to plan a path to navigate in unknown environments for exploration purposes. To the best of our knowledge, there is no such research developing a geometric map using a WiFi signal without relying on crowdsourcing or exteroceptive sensors such as a camera, radar, or lidar.

In the future, the work should explore machine learning methods that can improve the quality of the map obtained in a post-processing step. The integration of WiFi-based localization methods, which provide highly accurate trajectories, should also be integrated with the WiFi-based mapping method presented in this work to create an entirely independent end-to-end WiFiSLAM system. Actively planning a trajectory to improve the map quality will be pursued. 

\section*{Acknowledgements}

We would like to thank Matthew Lisondra and Ishaan Mehta for their feedback and help in data acquisition.

\bibliographystyle{IEEEtran}
\bibliography{bibliography.bib}

\begin{thebibliography}{10}
\providecommand{\url}[1]{#1}
\csname url@samestyle\endcsname
\providecommand{\newblock}{\relax}
\providecommand{\bibinfo}[2]{#2}
\providecommand{\BIBentrySTDinterwordspacing}{\spaceskip=0pt\relax}
\providecommand{\BIBentryALTinterwordstretchfactor}{4}
\providecommand{\BIBentryALTinterwordspacing}{\spaceskip=\fontdimen2\font plus
\BIBentryALTinterwordstretchfactor\fontdimen3\font minus
  \fontdimen4\font\relax}
\providecommand{\BIBforeignlanguage}[2]{{%
\expandafter\ifx\csname l@#1\endcsname\relax
\typeout{** WARNING: IEEEtran.bst: No hyphenation pattern has been}%
\typeout{** loaded for the language `#1'. Using the pattern for}%
\typeout{** the default language instead.}%
\else
\language=\csname l@#1\endcsname
\fi
#2}}
\providecommand{\BIBdecl}{\relax}
\BIBdecl

\bibitem{thrun2002robotic}
S.~Thrun, ``Robotic mapping: A survey,'' \emph{Exploring Artificial
  Intelligence in the New Millennium}, p. 1–35, 2003.

\bibitem{kudo2017utilizing}
T.~Kudo and J.~Miura, ``{Utilizing WiFi signals for improving SLAM and person
  localization},'' in \emph{{2017 IEEE/SICE International Symposium on System
  Integration (SII)}}.\hskip 1em plus 0.5em minus 0.4em\relax IEEE, 2017, pp.
  487--493.

\bibitem{ferris2007wifi}
B.~Ferris, D.~Fox, and N.~D. Lawrence, ``{WiFi-SLAM Using Gaussian Process
  Latent Variable Models},'' in \emph{{IJCAI}}, vol.~7, no.~1, 2007, pp.
  2480--2485.

\bibitem{zhou2021crowdsourcing}
B.~Zhou, W.~Ma, Q.~Li, N.~El-Sheimy, Q.~Mao, Y.~Li, F.~Gu, L.~Huang, and
  J.~Zhu, ``{Crowdsourcing-based indoor mapping using smartphones: A survey},''
  \emph{ISPRS Journal of Photogrammetry and Remote Sensing}, vol. 177, pp.
  131--146, 2021.

\bibitem{arun2022p2slam}
A.~Arun, R.~Ayyalasomayajula, W.~Hunter, and D.~Bharadia, ``{P2SLAM: Bearing
  based WiFi SLAM for Indoor Robots}.''

\bibitem{o1987art}
J.~O'rourke \emph{et~al.}, \emph{{Art gallery theorems and algorithms}}.\hskip
  1em plus 0.5em minus 0.4em\relax Oxford University Press Oxford, 1987,
  vol.~57.

\bibitem{xiong2017diversified}
H.~Xiong and D.~Tao, ``{A diversified generative latent variable model for
  wifi-slam},'' in \emph{{Proceedings of the AAAI Conference on Artificial
  Intelligence}}, vol.~31, no.~1, 2017.

\bibitem{miyagusuku2016improving}
R.~Miyagusuku, A.~Yamashita, and H.~Asama, ``{Improving gaussian processes
  based mapping of wireless signals using path loss models},'' in \emph{{2016
  IEEE/RSJ International Conference on Intelligent Robots and Systems
  (IROS)}}.\hskip 1em plus 0.5em minus 0.4em\relax IEEE, 2016, pp. 4610--4615.

\bibitem{huang2011efficient}
J.~Huang, D.~Millman, M.~Quigley, D.~Stavens, S.~Thrun, and A.~Aggarwal,
  ``{Efficient, generalized indoor WiFi GraphSLAM},'' in \emph{{2011 IEEE
  international conference on robotics and automation}}.\hskip 1em plus 0.5em
  minus 0.4em\relax IEEE, 2011, pp. 1038--1043.

\bibitem{liu2019collaborative}
R.~Liu, S.~H. Marakkalage, M.~Padmal, T.~Shaganan, C.~Yuen, Y.~L. Guan, and
  U.-X. Tan, ``{Collaborative SLAM based on WiFi fingerprint similarity and
  motion information},'' \emph{IEEE Internet of Things Journal}, vol.~7, no.~3,
  pp. 1826--1840, 2019.

\bibitem{herranz2016wifi}
F.~Herranz, {\'A}.~Llamazares, E.~Molinos, M.~Oca{\~n}a, and M.~Sotelo, ``{WiFi
  SLAM algorithms: An experimental comparison},'' \emph{Robotica}, vol.~34,
  no.~4, p. 837, 2016.

\bibitem{he2015wi}
S.~He and S.-H.~G. Chan, ``{Wi-Fi Fingerprint-Based Indoor Positioning: Recent
  Advances and Comparisons},'' \emph{IEEE Communications Surveys \& Tutorials},
  vol.~18, no.~1, pp. 466--490, 2015.

\bibitem{luo2014piloc}
C.~Luo, H.~Hong, and M.~C. Chan, ``{PiLoc: A self-calibrating participatory
  indoor localization system},'' in \emph{{International Symposium on
  Information Processing in Sensor Networks}}, pp. 143--153.

\bibitem{shin2011unsupervised}
H.~Shin, Y.~Chon, and H.~Cha, ``{Unsupervised construction of an indoor floor
  plan using a smartphone},'' \emph{IEEE Transactions on Systems, Man, and
  Cybernetics, Part C (Applications and Reviews)}, vol.~42, no.~6, pp.
  889--898, 2011.

\bibitem{zhou2015alimc}
B.~Zhou, Q.~Li, Q.~Mao, W.~Tu, X.~Zhang, and L.~Chen, ``{ALIMC: Activity
  landmark-based indoor mapping via crowdsourcing},'' \emph{IEEE Transactions
  on Intelligent Transportation Systems}, vol.~16, no.~5, pp. 2774--2785, 2015.

\bibitem{zhou2018graph}
B.~Zhou, Q.~Li, G.~Zhai, Q.~Mao, J.~Yang, W.~Tu, W.~Xue, and L.~Chen, ``{A
  graph optimization-based indoor map construction method via crowdsourcing},''
  \emph{IEEE Access}, vol.~6, pp. 33\,692--33\,701, 2018.

\bibitem{shen2013walkie}
``{Walkie-Markie: Indoor pathway mapping made easy}, author={Shen, Guobin and
  Chen, Zhuo and Zhang, Peichao and Moscibroda, Thomas and Zhang, Yongguang},''
  in \emph{{Symposium on Networked Systems Design and Implementation}}, 2013,
  pp. 85--98.

\bibitem{jiang2013hallway}
Y.~Jiang, Y.~Xiang, X.~Pan, K.~Li, Q.~Lv, R.~P. Dick, L.~Shang, and
  M.~Hannigan, ``{Hallway based automatic indoor floorplan construction using
  room fingerprints},'' in \emph{{ACM international joint conference on
  Pervasive and ubiquitous computing}}, 2013, pp. 315--324.

\bibitem{alzantot2012crowdinside}
{Alzantot, Moustafa and Youssef, Moustafa}, ``{CrowdInside: automatic
  construction of indoor floorplans},'' in \emph{{Proceedings of the 20th
  International Conference on Advances in Geographic Information Systems}},
  2012, pp. 99--108.

\bibitem{liang2016sensewit}
J.~Liang, Y.~He, and Y.~Liu, ``{SenseWit: Pervasive floorplan generation based
  on only inertial sensing},'' in \emph{{International Conference on
  Distributed Computing in Sensor Systems}}.\hskip 1em plus 0.5em minus
  0.4em\relax IEEE, 2016, pp. 1--8.

\bibitem{gonzalez2013cooperative}
A.~Gonzalez-Ruiz and Y.~Mostofi, ``{Cooperative robotic structure mapping using
  wireless measurements—A comparison of random and coordinated sampling
  patterns},'' \emph{IEEE Sensors Journal}, vol.~13, no.~7, pp. 2571--2580,
  2013.

\bibitem{deissler2010uwb}
T.~Deissler and J.~Thielecke, ``{UWB-SLAM with Rao-Blackwellized Monte Carlo
  data association},'' in \emph{{2010 International Conference on Indoor
  Positioning and Indoor Navigation}}.\hskip 1em plus 0.5em minus 0.4em\relax
  IEEE, 2010, pp. 1--5.

\bibitem{deissler2012infrastructureless}
T.~Dei{\ss}ler, M.~Janson, R.~Zetik, and J.~Thielecke, ``{Infrastructureless
  indoor mapping using a mobile antenna array},'' in \emph{{International
  Conference on Systems, Signals and Image Processing}}.\hskip 1em plus 0.5em
  minus 0.4em\relax IEEE, 2012, pp. 36--39.

\bibitem{fabila2009modem}
R.~Fabila-Monroy, A.~R. Vargas, and J.~Urrutia, ``{On modem illumination
  problems},'' \emph{XIII encuentros de geometria computacional, Zaragoza,
  Spain}, 2009.

\bibitem{klee1969every}
V.~Klee, ``{Is every polygonal region illuminable from some point?}'' \emph{The
  American Mathematical Monthly}, vol.~76, no.~2, pp. 180--180, 1969.

\bibitem{bajuelos2012hybrid}
A.~L. Bajuelos, S.~Canales, G.~Hern{\'a}ndez-Penalver, and A.~M. Martins, ``{A
  Hybrid Metaheuristic Strategy for Covering with Wireless Devices},'' \emph{J.
  UCS}, vol.~18, no.~14, pp. 1906--1932, 2012.

\bibitem{bahoo2020computing}
Y.~Bahoo, P.~Bose, S.~Durocher, and T.~C. Shermer, ``{Computing the
  k-Visibility Region of a Point in a Polygon},'' \emph{Theory of Computing
  Systems}, vol.~64, no.~7, pp. 1292--1306, 2020.

\bibitem{bahoo2019time}
Y.~Bahoo, B.~Banyassady, P.~K. Bose, S.~Durocher, and W.~Mulzer, ``{A
  time--space trade-off for computing the k-visibility region of a point in a
  polygon},'' \emph{Theoretical Computer Science}, vol. 789, pp. 13--21, 2019.

\bibitem{fafoutis2015rssi}
X.~Fafoutis, E.~Mellios, N.~Twomey, T.~Diethe, G.~Hilton, and R.~Piechocki,
  ``{An RSSI-based wall prediction model for residential floor map
  construction},'' in \emph{{World Forum on Internet of Things}}.\hskip 1em
  plus 0.5em minus 0.4em\relax IEEE, 2015, pp. 357--362.

\bibitem{"li2019houseexpo"}
L.~Tingguang, H.~Danny, L.~Chenming, Z.~Delong, W.~Chaoqun, and M.~Q.-H. Meng,
  ``{HouseExpo: A Large-scale 2D Indoor Layout Dataset for Learning-based
  Algorithms on Mobile Robots},'' \emph{arXiv preprint arXiv:1903.09845}, 2019.

\bibitem{grisetti2007improved}
G.~Grisetti, C.~Stachniss, and W.~Burgard, ``Improved techniques for grid
  mapping with rao-blackwellized particle filters,'' \emph{IEEE transactions on
  Robotics}, vol.~23, no.~1, pp. 34--46, 2007.

\end{thebibliography}

\end{document}